\documentclass[twoside,11pt]{article}

\usepackage{jmlr2e}
 \usepackage{booktabs}
\usepackage{hyperref}
\hypersetup{
    colorlinks=true,
    linkcolor=blue,
    filecolor=magenta,      
    urlcolor=cyan,
}
\usepackage{listings}
\usepackage{color}
\usepackage{MnSymbol}
\usepackage{wasysym}
\definecolor{dkgreen}{rgb}{0,0.6,0}
\definecolor{gray}{rgb}{0.5,0.5,0.5}
\definecolor{mauve}{rgb}{0.58,0,0.82}
\firstpageno{1}

\begin{document}

\title{PyODDS: An End-to-End Outlier Detection System}

\author{\name Yuening Li \email yueningl@tamu.edu \\
       \name Daochen Zha \email daochen.zha@tamu.edu \\
       \addr Department of Computer Science and Engineering \\
       \name Na Zou \email nzou1@tamu.edu\\
       \addr Department of Industrial \& Systems Engineering\\
       \name Xia Hu \email xiahu@tamu.edu \\
       \addr Department of Computer Science and Engineering\\
       Texas A\&M University\\
       College Station, TX 77840, USA}

\editor{}

\maketitle

\begin{abstract}
\textbf{PyODDS} is an end-to-end \textbf{Py}thon system for \textbf{O}utlier \textbf{D}etection with \textbf{D}atabase \textbf{S}upport. It provides various outlier detection algorithms which meet the demands for users in different fields, with or without data science or machine learning background. PyODDS gives the ability to execute machine learning algorithms in-database without moving data out of the database server or over the network. It also provides access to a wide range of outlier detection algorithms, including statistical analysis and more recent deep learning based approaches. \textbf{PyODDS} is released under the MIT open-source license, and publicly available at \url{https://github.com/datamllab/pyodds} with official documentations at \url{https://pyodds.github.io/}.
\end{abstract}


\section{Introduction}
Outliers refer to the objects with patterns or behaviors that are significantly rare and different with the rest of majorities. Outlier detection plays an important role in various applications, such as fraud detection, cyber security, medical diagnosis and industrial manufacturer. Overtime, a variety of anomaly detection approaches have been specifically developed for certain application domains. 

Recently, efforts have been made to integrate various anomaly detection algorithms into a single package with unified interfaces. Existing approaches~\cite{Constantinou2018,achtert2010visual,hofmann2013rapidminer,zhao2019pyod} contain different outlier detection methods for static data with various programming languages,  yet they do not tackle with time series data, and do not cater specifically to backend-servers. 

Time series data, as a series of data points indexed in time order, is ubiquitous in many real-world applications. For example, in transaction logs, instances usually have temporal correlations with time-stamps, such as unexpected spikes, drops, trend changes and level shifts. Despite its generability, there is no available outlier detection system that can efficiently and simultaneously deal with static data and time series data.

It is a challenging task to develop the above unified outlier detection system. First, the burden of throughput is heavier than usual since time window based approaches require to access and query the data in a more frequent way. Second, loading data from remote servers causes a large cost of moving data outside database server or over the network for analysis. Third, the system should support static and time series data with unified API.

We present \textbf{PyODDS}, a full stack, end-to-end system for outlier detection, which supports both static and time series data. $\textbf{PyODDS}$ has desirable features from the following perspectives. First, it contains 13 algorithms, including statistical approaches, and recent neural network frameworks. Second, PyODDS supports both static and time series data analysis, with flexible time-slices segmentation. Third,  PyODDS supports operation and maintenance from a light-weight SQL based database, which reduces the cost of queries and loading data from different remote servers. Fourth, PyODDS provides visualization tools for the original distribution of raw data, and predicted results, which offers users a direct and vivid perception. Last, PyODDS includes a unified API with detailed documentation, such as outlier detection approaches, database operations, and visualization functions.





\begin{figure}[t]
  \includegraphics[width=0.9\linewidth]{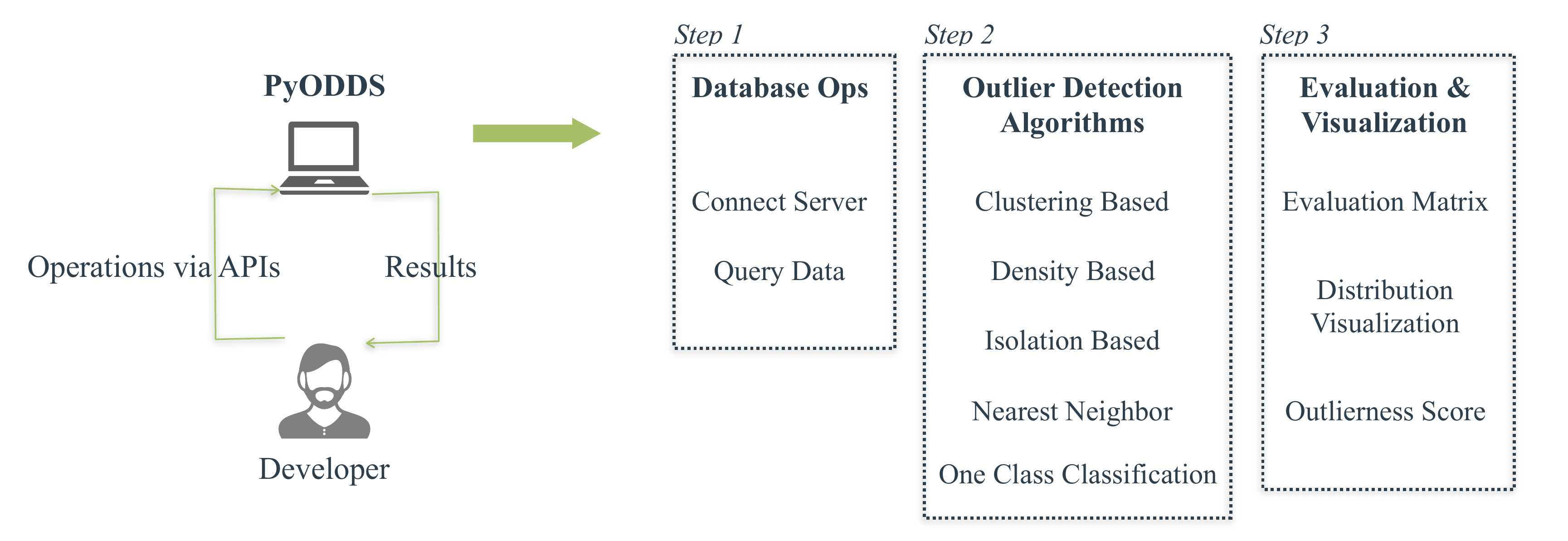}
  \caption{Overview of PyODDS}
  \label{pipeline}
  \vspace{-10pt}
\end{figure}

\section{PyODDS System}

Our system is written in Python and uses TDengine as the database support service. The pipline from query data to evaluation is outlined in Figure ~\ref{pipeline}. PyODDS follows the API design of scikit-learn All implemented methods (as shown in Table~\ref{algorithm})  are formulated as individual classes with same interfaces: (1) $fit$ function is to fit the selected model according to the given training data; (2) $predict$ function returns a binary class label corresponding to each instance in testing sets; (3) $decision\_function$ produces an outlier score for each instance to denote their outliernesses.

PyODDS also includes database operation functions for client users: (1) $connect\_server$ function allows the client  to connect the server with host address and user information; (2) $query\_data$ returns a pandas DataFrame containing time series retrieved from a given time range. For other server-side database operations, they are supported by the backend database platform TDengine, which provides caching, stream computing, message queuing and other functionalities.

\begin{table}[t]
\centering
\scriptsize
\begin{tabular}{@{}lllc@{}}

\toprule
\textbf{Methods}   &   \textbf{Reference}    & \textbf{Class API}   & \textbf{Category} \\ 
\midrule
CBLOF  &   ~\cite{he2003discovering}    & algo.cblof.CBLOF  & Fixed-length, shallow \\
SOD   & ~\cite{kriegel2009outlier} & algo.sod.SOD & Fixed-length, shallow\\
HBOS  &    ~\cite{aggarwal2015outlier}      & algo.hbos.HBOS      & Fixed-length, shallow    \\
IFOREST   &    ~\cite{liu2008isolation}      & algo.iforest.IFOREST  & Fixed-length, shallow \\
KNN  &    ~\cite{ramaswamy2000efficient}    & algo.knn.KNN   & Fixed-length, shallow \\
LOF &   ~\cite{breunig2000lof}   & algo.cblof.CBLOF & Fixed-length, shallow\\
OCSVM  &  ~\cite{scholkopf2001estimating}   & algo.ocsvm.OCSVM  & Fixed-length, shallow \\
PCA    &   ~\cite{shyu2003novel}  & algo.pca.PCA & Fixed-length, shallow \\
AUTOENCODER   &  ~\cite{hawkins2002outlier}  & \begin{tabular}{@{}c@{}}algo.autoencoder \\ .AUTOENCODER\end{tabular} & Fixed-length, deep \\
DAGMM & ~\cite{zong2018deep} & algo.dagmm.DAGMM  & Fixed-length, deep  \\
LSTMENCDEC   &  ~\cite{malhotra2016lstm}    & \begin{tabular}{@{}c@{}}algo.lstm\_enc\_dec\_axl \\ .LSTMED\end{tabular}   & Time series, deep\\
LSTMAD    &   ~\cite{malhotra2015long}     & algo.lstm\_ad.LSTMAD  & Time series, deep    \\
LUMINOL    &    & algo.luminol.LUMINOL   &   Time series, shallow          \\ \bottomrule
\end{tabular}
\caption{Outlier detection models in PyODDS}
\label{algorithm}
\end{table}

\noindent\begin{minipage}{\textwidth}
\scriptsize
\begin{lstlisting}[caption={Demo of PyODDS API},captionpos=b, label=demo]
>>> from utils.import_algorithm import algorithm_selection
>>> from utils.utilities import  output_performance,connect_server,query_data

>>> # connect to the database
>>> conn,cursor=connect_server(host, user, password)

>>> # query data from specific time range
>>> data = query_data(database_name,table_name,start_time,end_time)

>>> # train the anomaly detection algorithm
>>> clf = algorithm_selection(algorithm_name)
>>> clf.fit(X_train)

>>> # get outlier result and scores
>>> prediction_result = clf.predict(X_test)
>>> outlierness_score = clf.decision_function(X_test)
==============================
Results in Algorithm dagmm are:
accuracy_score: 0.98
precision_score: 0.99
recall_score: 0.99
f1_score: 0.99
roc_auc_score: 0.99
processing time: 15.330137 seconds
==============================
>>> # evaluate and visualize the prediction_result
>>> output_performance(X_test,prediction_result,outlierness_score)
>>> visualize_distribution(X_test,prediction_result,outlierness_score)
\end{lstlisting}
\end{minipage}

Moreover, PyODDS includes a set of utility functions for model evaluation and visualization: (1) $visualize\_distribution$ is to visualize the original distribution for static data and time series; (2) $visualize\_outlierscore$ is to visualize the predicted outlier score in testing cases; (3) $output\_performance$ produces the evaluation for the performance of the given algorithm, including accuracy score, precision score, recall score, f1 score, roc-auc score and processing time cost.

\begin{figure}
  \includegraphics[width=\linewidth]{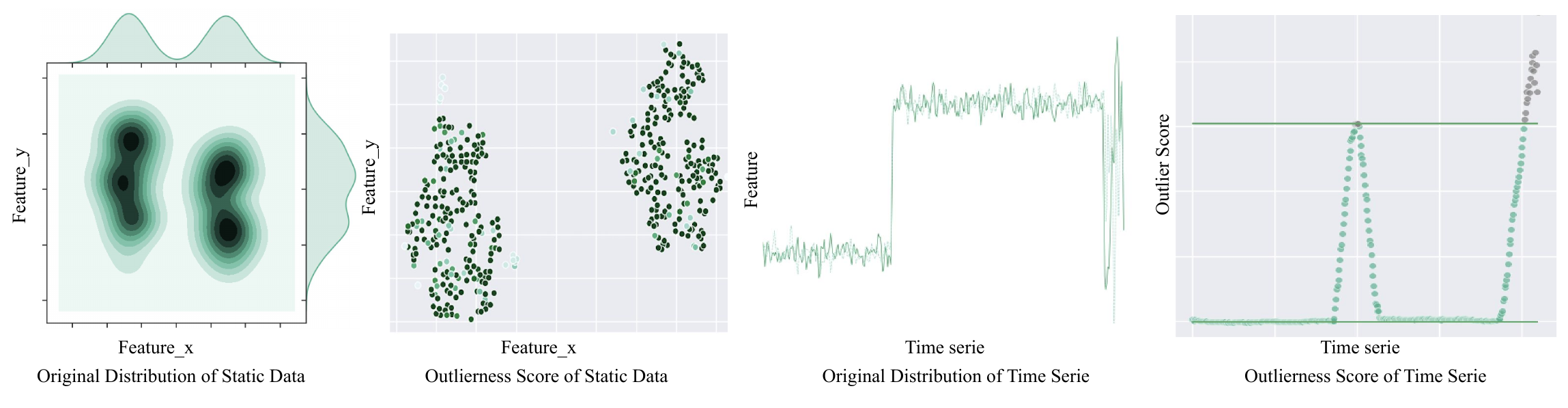}
  \caption{Demonstration of using PyODDS in visualizing prediction result}
  \label{visualization}
\end{figure}

A PyODDS API demo is shown in Listing~\ref{demo}. Lines 1-3 import the utility functions. Lines 4-5 create a connection object $conn$ with cursor  $cursor$ connecting to the dataset by the host address and user information. Lines 7-8 show how to query data from a given time range. Line 10-12 declare an object $clf$ as a specific algorithm for data analysis, and fit the model through the training data. Lines 14-16 produce the predicted result for the testing data. Lines 18-20 give a visualization of the prediction results. 
Examples of visualization functions are shown in Figure ~\ref{visualization}. The two-dimensional artificial data used in the example is created by hand-crafted data where inliers from two Gaussian distribution and outliers as random noises from other distribution. The first and the second figure (from left to right) denote the distribution of static data: the first one uses the kernel density estimation procedure to visualize the original distribution, and the second one plots the instances as scatters where lighter shades denote to higher outlierness score.  The third and fourth figure visualize the prediction results for time series: the third one plots original features as curves where the x axis denotes timestamp, y axis denotes the features, the fourth figure shows the outlierness score in y axis, corresponding to the timestamp in x axis. 


\section{Conclusion and Future Work}
This paper introduces $\textbf{PyODDS}$, an open-source  system for anomaly detection  utilizing state-of-the-art machine learning techniques.   It contains 13 algorithms, including classical statistical approaches, and recent neural network frameworks for static and time series data. It also provides an end-to-end solution for individuals as well as enterprises,  and supports operations and maintenance from light-weight SQL based database to back-end machine learning algorithms. As a full-stack system, PyODDS aims to lower the threshold of learning scientific algorithms and reduces the skills requirement from both database and machine learning sides. In the future, we plan to enhance the system by implementing models for heterogeneous data~\cite{li2019specae,huang2019graph,li2019deepstruc}, improving the interpretability and reliability of the algorithms~\cite{liu2019single}, and integrating more advanced outlier detection methods.

\bibliographystyle{ieeetr}
\bibliography{sample}

\end{document}